\documentclass{article}

\usepackage[preprint]{packages/neurips_2025}

\usepackage[utf8]{inputenc} 
\usepackage[T1]{fontenc}    
\usepackage[backref=page]{hyperref}       
\usepackage{xurl}            
\usepackage{booktabs}       
\usepackage{amsfonts}       
\usepackage{nicefrac}       
\usepackage[stretch=10,shrink=10,step=1,protrusion=true,expansion=true,final]{microtype}
\usepackage[x11names]{xcolor}       
\usepackage{tikz}
\usetikzlibrary{bayesnet}

\title{Extending Epistemic Uncertainty Beyond Parameters Would Assist in Designing Reliable LLMs}

\author{%
  T. Duy Nguyen-Hien\\
  Department of Computer Science\\
  School of Computing\\
  National University of Singapore\\
  \texttt{duynht@comp.nus.edu.sg}\\
  \And
  Desi R. Ivanova\\
  Department of Statistics\\
  University of Oxford\\
  \texttt{desi.ivanova@stats.ox.ac.uk}\\
  \And
  Yee Whye Teh\\
  Department of Statistics\\
  University of Oxford\\
  \texttt{y.w.teh@stats.ox.ac.uk}\\
  \And
  Wee Sun Lee\\  
  Department of Computer Science\\
  School of Computing\\
  National University of Singapore\\
  \texttt{leews@comp.nus.edu.sg}
}

\usepackage{amsmath}
\usepackage{amssymb}
\usepackage{mathtools}
\usepackage{amsthm}
\usepackage{bm}
\usepackage[capitalize,noabbrev]{cleveref}
\usepackage[edges]{forest}
\usepackage{xfrac}

\theoremstyle{plain}
\newtheorem{theorem}{Theorem}[section]
\newtheorem{proposition}[theorem]{Proposition}

\theoremstyle{definition}

\theoremstyle{remark}

\usepackage[textsize=tiny]{todonotes}

\usepackage{fontawesome}
\usepackage{soul}
\usepackage{multirow}

\begin{document}

\maketitle

\begin{abstract}
        Although large language models (LLMs) are highly interactive and extendable, current approaches to ensure reliability in deployments remain mostly limited to rejecting outputs with high uncertainty in order to avoid misinformation. This conservative strategy reflects the current lack of tools to systematically distinguish and respond to different sources of uncertainty. In this paper, \textbf{we advocate for the adoption of Bayesian Modeling of Experiments -- a framework that provides a coherent foundation to reason about uncertainty and clarify the reducibility of uncertainty -- for managing and proactively addressing uncertainty that arises in LLM deployments}. This framework enables LLMs and their users to take contextually appropriate steps, such as requesting clarification, retrieving external information, or refining inputs. By supporting active resolution rather than passive avoidance, it opens the door to more reliable, transparent, and broadly applicable LLM systems, particularly in high-stakes, real-world settings.

\end{abstract}

\section{Advancing the Reliability of Large Language Models Calls for Addressing Uncertainty}\label{sec:intro}

    Large language models (LLMs) are machine learning systems trained on a very large corpus of text for the conditional generation of textual tokens. More precisely, they are optimized to predict the next token given a sequence of preceding tokens by modeling a predictive distribution $P(X_t|X_{<t})$. Due to the combined increase in scale of training corpora, model architectures, and computation power, modern large language models are not only able to generate coherent text, but also exhibit emergent capabilities such as common sense knowledge, test-time task-adaptation, problem-solving, and programming~\citep{brown_language_2020, chen_evaluating_2021, wei_emergent_2022, li_starcoder_2023}. These capabilities enable LLMs to become general processors that are interactive and integrative via textual interfaces.

    On one hand, there is a growing body of research looking to investigate the uncertainty in LLM generation, with much focus on measuring total uncertainty~\citep[\textit{inter alia}]{lin_teaching_2022, tian_just_2023, xiong_can_2024, kadavath_language_2022, kuhn_semantic_2023, farquhar_detecting_2024, yadkori_mitigating_2024} towards reliable widespread deployments of LLMs. Most of these proposed uncertainty estimates are evaluated for their calibration with respect to the true correctness of responses, implying the sole utility of accepting or rejecting responses (may entail deferral instead of sole rejection~\citep{gupta_language_2024}). This is motivated as a hedge against a prominent issue of state-of-the-art LLMs -- the risk of generating misinformation, often known as \textit{hallucination}~\citep{farquhar_detecting_2024}. Intuitively, the hallucination risk correlates with total uncertainty in the predictive distribution -- higher variance increases the likelihood of selecting low-probability, potentially incorrect outputs during test-time sampling. The easiest way to mitigate the risk of hallucination is by acting conservatively and abstaining from generation, thereby avoiding being wrong.

    On the other hand, much fewer papers have sought to actively reduce uncertainty, or even just decompose the total uncertainty~\citep{zhang_clarify_2023, hou_decomposing_2024, park_clara_2024, kobalczyk_active_2025, ling_uncertainty_2024}. While frontier LLMs have been consistently improving on benchmarks~\citep[\textit{inter alia}]{srivastava_beyond_2023, chiang_chatbot_2024}, including improvements on calibrations~\citep{phan_humanitys_2025}, their practical reliability is often contingent on users issuing detailed and well-structured prompts~\citep{zamfirescu-pereira_why_2023, carlini_yet_2024, kobalczyk_active_2025, vijayvargiya_interactive_2025}. For example, a simple request such as “Write a post announcing a new post-doc position” may yield responses that vary in length, are too vague, overly formal, or insufficiently engaging. Often, users need a more specific prompt like “Write a short (approximately 250 words) compelling LinkedIn post to attract recent PhD graduates to a new post-doc opening in our AI research group at XYZ University. Target audience with a background in K. Include a link to the full posting, mention the application deadline, and use an approachable, enthusiastic tone to encourage sharing” to produce a desirable result. \textit{This need for specificity exhibits that LLMs are subject to uncertainty from ambiguities in human communication, due to the inherent ambiguity of natural language}. Qualities such as \textit{polysemy}, where words or phrases have multiple interpretations, can introduce significant variability in model outputs due to multiplicities of possible interpretations. Since a good interactive system should reduce users' thinking effort~\citep{krug_dont_2013}, requiring users to precisely articulate every aspect of their intent undermines usability and limits adoption. Therefore, addressing uncertainty due to ambiguity in human–LLM interaction is essential to building more reliable AI systems.
    
    Further, among the most ambitious applications, LLMs are envisioned  as the central reasoning engines for autonomous agents -- systems capable of \textit{experiencing} their environment, \textit{reasoning} over observations, and \textit{interacting} with open-ended environments~\citep{lehman_evolution_2025, liu_advances_2025}. LLM-based agents typically operate through textual interfaces, such as Application Programming Interfaces (API) requests and responses, text editors, and command-line interfaces (shells), to perform tasks like information retrieval, programming, or interacting with external tools. A notable example is the \textit{AI Scientist} project~\citep{lu_ai_2024}, which aims to enable LLMs to conduct scientific research independently. \textit{Much like human researchers, such systems must reduce the uncertainty about the open-ended world by various procedures: gathering knowledge, synthesizing hypotheses from knowledge and observations, designing experiments, and verifying hypotheses.}

    Evidently, progress in the reliability of large language models increasingly depends on our ability to identify, characterize, and systematically reduce the uncertainty that arises from various sources during their deployment. While improvements in scale and architecture have yielded powerful general-purpose models, these advances alone are insufficient for ensuring reliability in real-world applications. As LLMs are tasked with open-ended reasoning, complex interaction, and autonomous decision-making, uncertainty -- whether stemming from ambiguous inputs, incomplete knowledge, or model limitations -- becomes a central obstacle. Addressing these uncertainties is not a peripheral concern, but a foundational requirement for the next stage of LLM development. Rather than merely abstaining from action in uncertain scenarios, future systems must be equipped to recognize and proactively rectify such uncertainty.
    
    In this paper, \textbf{we argue that the conventional perspective about epistemic uncertainty in machine learning -- typically focused on uncertainty in model weights -- should be expanded to fully study and address new challenges. We suggest the Bayesian Modeling of Experiments framework~\citep{lindley_measure_1956, degroot_uncertainty_1962, rainforth_modern_2024} as a coherent and suitable approach that could help tackle these emerging challenges}. Bayesian Modeling of Experiments not only elucidates the reducibility of many types of uncertainty encountered in the deployment of LLMs but also establishes a coherent foundation, enabling users and developers to more effectively understand model behaviors. Crucially, it moves beyond treating uncertainty as a reason for inaction and instead supports mechanisms for actively resolving uncertainty.

    \textbf{Organization~~~} In Section~\ref{sec:prelim}, we present the foundation of uncertainty quantification and the Bayesian Modeling of Experiments framework. We show in Section~\ref{sec:retro-llm} that BME is methodologically consistent with many new impromptu approaches in reducing uncertainty and, in fact, unifies them under the same line of reasoning. Section~\ref{sec:future} discusses the potential of BME in coherently extending current approaches. Finally, we draw conclusions and acknowledge opposing views in Section~\ref{sec:conclusion}.
    
\section{Uncertainty Quantification and Bayesian Modeling of Experiments}\label{sec:prelim}
    \subsection{Foundations}
        
        The primary objective of uncertainty quantification is to evaluate a model's confidence in its predictions, thereby enabling more informed decision-making and effective risk mitigation. The nature of uncertainties and methods to address them has been a persistent topic of discussion in modern science, by statisticians, scientists, engineers, and other specialists~\citep{hora_aleatory_1996, kiureghian_aleatory_2009, fox_distinguishing_2011}. 
        
        While there can be many sources of uncertainty, the canonical categorization for the character of uncertainties are the two types: aleatoric uncertainty and epistemic uncertainty. The word ``aleatoric'' is derived from the Latin \textit{alea}, meaning the rolling of dice. It characterizes types of uncertainties that are presumably irreducible and intrinsic to the randomness of a phenomenon. ``Epistemic'', on the other hand, roots in the Greek word \textit{episteme}, meaning pertaining to knowledge. It is thus used to label uncertainties that are caused by the lack of knowledge and presumably reducible with more complete knowledge. This classification, as it turns out, is more of a convenient means of communication in the realm of modeling~\citep{kiureghian_aleatory_2009}. For example, while a coin toss is normally considered aleatoric in nature, it is arguable that once all the physical states of the coin is fully known, one can predict which face it lands on with higher confidence than chance. In such a case, the aleatoric uncertainty has morphed into epistemic uncertainty. This illustrates that the distinguishability between \textit{epistemic uncertainty} and \textit{aleatoric uncertainty} is only meaningful and unambiguous within the confines of a model of analysis. In one model, a source of uncertainty may be aleatoric, in another, it may be epistemic. 

    \subsection{Representing Reducible Uncertainty via Bayesian Modeling of Experiments and Shannon Information}
        The Bayesian modeling of experiment analysis~\citep{degroot_uncertainty_1962, lindley_measure_1956}, which we refer to as Bayesian Modeling of Experiments (BME), considers situations in which it is desired to gain knowledge about the true value of some parameters (or more generally, about a true state of the world) by means of experimentation, thereby reducing uncertainty. In the simplest setting, an experiment is represented by a random variable $\Lambda$ defined on some probability space, and performing an experiment is the same as observing a realization $\lambda$ of $\Lambda$. 
        
        Let $\Omega$ denote the set of all possible values of the parameter of interest $\xi$. The random variable $\Xi$ of this parameter and $\Lambda$ together specify a conditional probability distribution $P(\Lambda|\Xi)$ of outcomes $\lambda$ given each possible value of $\xi$. For example, suppose you are experimenting with different flavorings to find which one makes the most appealing cake to your family, then $\Omega = \{ \verb|vanilla, almond, lemon, coffee|\}$. The random variable $\Xi$ represents uncertainty over which flavoring is the best. For each choice of flavoring, your outcome is obtained from the panel of family members with ratings among \verb|delicious, bland,| or \verb|abhorrent|. 
        
        Suppose that the experimenter's knowledge about the true value of $\xi$ can be expressed, after each experiment, in terms of a probability distribution $\mu$ defined over $\Omega$. Each distribution $\mu$ indicates a certain amount of uncertainty on the part of the experimenter about the true value of $\xi$. It is assumed that for each $\mu$ this uncertainty can be characterized by a non-negative number, often chosen to be the Shannon entropy. 
        For simplicity, we consider discrete distributions throughout this paper. A probability distribution over $\Omega$ is a $|\Omega|$-dimensional vector $\mu = (\mu_1, \dots, \mu_k)$ in a $|\Omega|$-dimensional simplex -- i.e., $\sum_{i=1}^{|\Omega|} \mu_i = \sum_{\xi \in \Omega} \mu(\xi) = 1$ -- then its Shannon entropy is defined as

        \begin{equation}
            H[\mu] = -\sum_{i=1}^{|\Omega|} \mu_i \log \mu_i.
        \end{equation}

        Before the experiment, suppose our belief about $\xi$ is expressible via a prior $P(\Xi)$ over $\Omega$. The reduction of uncertainty after an experiment is, in turn, the difference between the uncertainty of the prior distribution and the uncertainty of the posterior distribution. Since Shannon entropy is often referred to as the \textit{information} associated with a distribution, this reduction in uncertainty is often referred to as the \textit{information gain} of the experiment. For each experiment, the pointwise information gain is 

        \begin{align}
            I[\Xi;\lambda]  &= H[P(\Xi)] - H[P(\Xi | \lambda)]\\
                            &= -\sum_{\xi} p(\xi) \log p(\xi)  + \sum_{\xi} p(\xi | \lambda) \log p(\xi | \lambda).
        \end{align}

        The expected information gain over the marginal $P(\Lambda) = \sum_{\xi}P(\Lambda|\xi)p(\xi)$ is correspondingly

        \begin{align}
            I[\Xi;\Lambda]  &= H[P(\Xi)] - \mathbb{E}_{P(\Lambda)} H[P(\Xi | \lambda)]\\
                            &= -\sum_{\xi} p(\xi) \log p(\xi)  + \sum_{\lambda} p(\lambda) \sum_{\xi} p(\xi | \lambda) \log p(\xi | \lambda)\\
                            &= \sum_{\lambda}\sum_{\xi} p(\xi,\lambda) \log p(\xi | \lambda) -\sum_{\xi} \sum_{\lambda} p(\xi, \lambda) \log p(\xi) \\
                            &= \sum_{\xi, \lambda} p(\xi, \lambda) \log \frac{p(\xi | \lambda)}{p(\xi)}\\
                            &= \sum_{\xi, \lambda} p(\xi, \lambda) \log \frac{p(\xi, \lambda)}{p(\xi) p(\lambda)}\\
                            &= D_{KL} (P(\Xi, \Lambda) \| P(\Xi) P(\Lambda))
        \end{align}

        This shows that the expected information gain coincides with the KL-divergence between the joint distributions and the product distribution of the parameter of interest and the experiment. This quantity is also referred to as the mutual information between $\Xi$ and $\Lambda$~\citep{cover_elements_2012}.
                
        In this paper, we distinguish BME from the more commonly known Bayesian Experimental Design~\citep[BED]{rainforth_modern_2024}, in that BED distinguishes an experiment $\lambda$ into its design $\lambda_d$ and its outcome $\lambda_o$. Continuing the baking example, different $\lambda_d$ is analogous to different family members from whom you obtain the ratings. This distinction made in \textit{BED is useful when you want to optimize for the best design} $\lambda_d$, such as in active learning (Appendix~\ref{sec:ml-retro}), while \textit{BME is a simpler and more general framework for quantification over random variables}.

        We demonstrate the soundness of BME in representing reducible uncertainty in Appendix~\ref{sec:ml-retro} by showing that it is compatible with conventional notions of uncertainty in supervised learning settings with fixed data and with adaptive data acquisition. \textbf{Under the unified view of BME, epistemic uncertainty is the uncertainty reducible via experimentation}. Thus, the distinction between epistemic uncertainty and (effective) aleatoric uncertainty depends on the experiment apparatus available.

\section{Moving Beyond Total Uncertainty and Rejection With BME}\label{sec:retro-llm}
    \subsection{Current Landscape of Uncertainty Quantification in LLMs}
      As mentioned in Section~\ref{sec:intro}, most research on uncertainty in LLM center on estimating total predictive uncertainty. \citet{lin_teaching_2022, tian_just_2023, xiong_can_2024} propose eliciting total predictive confidence measures by asking the LLM to generate numerical tokens directly or to perform auxiliary procedures in self-verified consistency of generations. These direct elicitation can be surprisingly correlated with task performance~\citep{kadavath_language_2022}. \citet{malinin_uncertainty_2021} propose estimating sequence-level predictive entropy by assuming Markov factorization of sequences over generated tokens. This approach is modified into semantic-level entropy by clustering sequences of equivalent semantics to overcome the redundancy in expressivity of natural languages~\citep{kuhn_semantic_2023, lin_contextualized_2024, aichberger_semantically_2024, nikitin_kernel_2024, lin_generating_2024}. \citet{kadavath_language_2022, kapoor_large_2024, kossen_semantic_2024} propose to supervised train uncertainty functions from the latent representations of LLMs' generations. \citet{yadkori_mitigating_2024} employ techniques in statistical risk control to propose a conformal abstention procedure that provides guarantees on hallucination rates. \citet{yang_bayesian_2024} approximate total predictive uncertainty by marginalizing over low-rank adaptation of model weights. Despite these advances, the prevailing focus on total uncertainty overlooks the need to disaggregate and target specific uncertainty sources -- an essential step for proactively mitigating uncertainty. 
        
    To our knowledge, only a handful of research studies, including ones discussed below~\citep[ \textit{inter alia}]{zhang_clarify_2023, hou_decomposing_2024, park_clara_2024, kobalczyk_active_2025}, consider the decomposition of uncertainty sources. However, most prior methods remain ad hoc in their conceptual treatment of uncertainty, limiting their potential impact. \textbf{By retrofitting these examples under the BME framework, we highlight the coherence reasoning process enabled by BME and its broad potential in addressing uncertainty in LLM deployments.}
    
    \subsection{Expected Information Gain in LLMs}
        For convenience in exposition, we look at the formal definition of a language model as a conditional probability distribution $P_{LM}(Y|X)$, where $Y$ and $X$ are sequence-level random variables (details in Appendix~\ref{sec:formal-lm}).

        Let $C$ represent an arbitrary auxiliary context provided in the prompt. Using the BME perspective, we can represent the uncertainty of $P_{LM}(Y|X)$ reducible with $C$ as the experiment stimulus as

        \begin{equation}
            I[Y|X;C] = H[P_{LM}[Y|X]] - \mathbb{E}_{P(C)} H[P_{LM}(Y|X, C)]
        \end{equation}

        Since $C$ is a sequence-valued random variable, it is extremely rich as it encapsulates different types of context, such as clarifications, prompt templates, or in-context demonstrations, etc. On top of that, LLMs are also powerful simulators for $P(C|X)$ in cases we want to optimize for the best context that is dependent on $X$.

        Note that instead of designating $Y$ to be a sequence-valued random variable, one can consider lifting the abstraction to the semantic level by considering $Y$ to be a semantic-valued random variable. Empirical distributions of the semantic $Y$ can then be aggregated from token-level and sequence-level probabilities using various approaches~\citep{kuhn_semantic_2023, farquhar_detecting_2024, lin_generating_2024, lin_contextualized_2024, aichberger_semantically_2024, nikitin_kernel_2024, hou_decomposing_2024, lamb_semantic-level_2025}. Such semantic probabilities remain compatible with the BME framework.

    \subsection{BME Elucidates the Mechanisms of Multiple Heuristics as Reducing Different Sources of Uncertainty}
    \subsubsection{Epistemic Intention Uncertainty: Quantify and reduce uncertainty caused by ambiguity in expressions of human intents}
        \citet{zhang_clarify_2023} are among the first to attempt to distinguish uncertainty due to lack of knowledge and uncertainty due to lack of clarity. They posit that in order to avoid tedious unnecessary clarification questions, ``systems [...] must predict a scalar uncertainty estimate $u(x)$ for each input $x$, that correlates with how much performance is expected to improve after clarification'', which aligns with the thesis of BME. Their proposed method \textsc{Intent-Sim}, however, diverges from this idea and employs \textit{semantic entropy of the self-clarified predictive posterior marginalized over simulated intents}. More specifically, they quantify $H[P(Y|x)] = H[\sum_cP(Y|x,c)P(c|x)]$. While their method shows improvement over methods that only approximate $P(Y|X) \approx P(Y|x, c=\varepsilon)$ (here $\varepsilon$ is the empty string),  it diverges from the intent to quantify uncertainty reducible via clarification, which should have been $I[Y|x; C|x] = I[Y;C|x] = H[P(Y|x)] - H[P(Y|x, C)]$.
        
        \citet{hou_decomposing_2024} arrive at the gap $H[P(Y|x)] - \mathbb{E}_{P(C|x)} H[P(Y|x,c)]$ with $C$ being clarifications of the question $x$ by adopting the reasoning of Bayesian neural networks~\citep{gal_dropout_2016} and demonstrate its superiority over total uncertainty $H[P(Y|x)]$ in detecting ambiguous questions. Interpreting under BME, their method \textit{quantifies} the reducible uncertainty ${H[P(Y|x)] - H[P(Y|x,C)]}$ by experimenting with different values of $C$, for each fixed $x$. Yet, \citet{hou_decomposing_2024} define the information gain as a quantification of aleatoric uncertainty due to its root in the data. They overlook the conflict by definition that aleatoric uncertainty is irreducible, while suggesting reducing this uncertainty by soliciting clarification from the user.
        
        In LLM-based robotics, \citet{park_clara_2024} recognize the need to distinguish ambiguous requests with infeasible requests. They elicit the uncertainty of the LLM by computing the entropy of the predictive posterior marginalized over contexts, similar to ~\citet{zhang_clarify_2023}, but with contexts being different scene description prompts. For each highly uncertainty request, they propose first verifying with the LLM the visibility, then letting the LLM arbitrarily ask the user for clarification of feasible requests while abstaining from infeasible ones. By utilizing the BME framework, it would be possible to further elicit the uncertainty due to ambiguity and to identify the most impactful question to ask the user by maximizing information gain.

        Among prior work, only \citet{kobalczyk_active_2025} employ the Bayesian Experimental Design framework. Yet, they only utilize it as an optimization procedure to select the most informative clarifying question by maximizing expected information gain $\max_{q} H[P(Y|x)] - \mathbb{E}_{a|q,x} H[Y|x,q,a]$, where $q$ is a clarifying question and $a$ is its corresponding answer from the user. They rely on a separate line of reasoning to conceptually distinguish between two types of uncertainty: one arising from model limitations (i.e., lack of knowledge) and the other from input ambiguity. This distinction depends on an externally defined notion of ambiguity, extrinsic to the LLM.
        We posit that the  BME framework allows for a unified treatment of these uncertainty sources, in terms of both quantification and resolution. For each question $x$, the uncertainty caused by ambiguity can by quantified as $I[Y;C|x] = H[P(Y|x)] - H[P(Y|x, C)]$, while the best clarification context can be selected by optimizing $\max_c H[P(Y|x)] - H[P(Y|x, c)]$, where $c$ could be concatenated clarifying question-answer pairs. The distinction between model-based and ambiguity-based uncertainty can emerge naturally from the experimental setup: by varying model parameters to probe model uncertainty, and varying input clarifications to probe ambiguity. This avoids the need for an external definition of ambiguity as in \citep{kobalczyk_active_2025} and instead treats it as model-relative, aligning with the intuitive notion that ambiguity is subjective to the interpreter -- in this case, the model.

    \subsubsection{Epistemic Presentation Uncertainty: Quantify and reduce uncertainty caused by prompt templates}
        One of the emergent capabilities of LLMs is test-time task-adaptation, commonly referred to as in-context learning (ICL). Simply by conditioning on a few demonstrations of the task in the prompt, LLMs can perform predictive tasks by means of conditional generation. Even more remarkably, they have achieved several state-of-the-art results, such as in multiple-choice question-answering, translation, and text classification~\citep{brown_language_2020}. The instruction and articulation of these few-shot demonstrations, often called \textit{prompt templates}, is important to the performance of LLMs on adapted tasks. The sensitivity of LLMs to semantic-invariant lexical transform, e..g, capitalization, can lead to high variance in performance~\citep{zhao_calibrate_2021, lu_fantastically_2022}. Many methods in selecting the best template are variants of maximizing ``mutual information between the input and the predicted output''~\citep{yang_improving_2024} that is first proposed in \citep{sorensen_information-theoretic_2022}. In their paper, \citet{sorensen_information-theoretic_2022} consider a set of $K$ different prompt templating transforms $\tau_i$ and quantify the quality of each template by the mutual information 

        \begin{align}
            I[\tau_i(X);Y] &= H[P(Y)] - H[P(Y|\tau_i(X))]\\
                           &\approx H\left[\frac{1}{N} \sum_{j=1}^{N} P(Y|x_j)\right] - \frac{1}{N}\sum_{j=1}^N H[P(Y|\tau_i(x_j))]
        \end{align}
        
        They hypothesize that templates with the highest mutual information improve performance and verify it empirically. 
        
        It turns out, in fact, $H\left[\frac{1}{N} \sum_{j=1}^{N} P(Y|x_j)\right] \approx H[P(Y|X)]$ (details in Appendix~\ref{sec:taylor-approx}), hence their computation is also an approximation of $H[P(Y|X)] - H[P(Y|X, \tau_i)]$. This makes the method by \citet{sorensen_information-theoretic_2022} interpretable under the BME framework, which can explain its effectiveness. Conceptually, \citet{sorensen_information-theoretic_2022} measures the reducible uncertainty caused by prompt templates, with the experiment apparatus being the set of different templates. Thus, the maximization of this approximation effectively reduces the uncertainty due to prompt templates ``confusing'' to the LLM, yielding performance gain. 

    \subsubsection{Epistemic Demonstration Uncertainty: Quantify and reduce uncertainty caused by in-context examples}

        Similar to prompt templates, selecting effective examples for few-shot demonstrations is also not trivial and significantly determines the performance. \citet{ling_uncertainty_2024} approach this problem through the lens of uncertainty, recognizing that ICL performance is sensitive to decoding configurations, as well as the set of few-shot examples.
        
        Yet, by the praxis of machine learning in attributing epistemic uncertainty to the model parameters and aleatoric uncertainty to the data, the formulation of uncertainties in \citep{ling_uncertainty_2024} yield incompatible peculiarities. In particular, they fix the decoding configuration and assume each set of few-shot examples associates with a latent concept parameter $z$. The \textit{epistemic uncertainty} under this modeling, nevertheless, is designated to be the uncertainty persisting across sets of demonstrations, particularly $\mathbb{E}_z H[P(Y|X,z)]$, even though the persistence should indicate its irreducibility under a fixed decoding configuration. The information gain from different sets of demonstrations $H[P(Y|X)] - \mathbb{E}_z H[P(Y|X,z)]$, on the other hand, is designated as \textit{aleatoric uncertainty} when the difference quantify the reduce of uncertainty by definition.
        
        From the view of BME, it is clear that our experiment apparatus is the set of few-shot examples, and the target quantification is $I[Y|X; Z] = H[P(Y|X)] - H[P(Y|X,Z)]$. The uncertainty associated with these examples is, hence, reducible via selecting examples that maximize the information gain. This insight provides additional opportunities to resolve this uncertainty beyond quantification -- i.e., by optimizing for $\max_z H[P(Y|x)] - H[P(Y|x, z)]$ for each question $x$. Furthermore, using different decoding configurations as the experiment apparatus could have elicited another separate source of reducible uncertainty.               

\section{A Future Research Agenda For Uncertainty in LLMs}\label{sec:future}
    \subsection{Handling Multiple Sources of Reducible Uncertainty}

        The BME framework generalizes the notion of epistemic uncertainty, offering a coherent and interpretable lens through which to study and manage uncertainty in LLMs. Beyond its conceptual clarity, this framework could be instrumental in improving LLM deployment by enabling the \textit{joint reduction} of multiple uncertainty sources, such as prompt ambiguity, template choice, or in-context examples, within a unified methodology.
        
        A natural but idealized strategy for uncertainty reduction would be to exhaustively optimize across all factors that influence the LLM output -- e.g., refining the prompt template, selecting the most informative examples, clarifying ambiguous inputs, and so on. However, this ``brute-force'' approach quickly becomes infeasible in real-world settings due to combinatorial complexity as more sources of uncertainty are recognized. 

        \newcommand{\psibayesnet}{\scalebox{0.6}{
            $\begin{array}{rcl} 
                                    &   A   &           \\ 
                        \nearrow    &       &   \searrow \\ 
                    X   \rightarrow &   B   &   \rightarrow \Psi \rightarrow Y \\ 
                        \searrow    &       &   \nearrow \\ 
                                    &   E   & 
            \end{array}$        
        }}
        
        To motivate more tractable solutions, consider an idealized case where we design a controllable proxy variable $\Psi$ that reflects the combined effects of multiple uncertainty sources: ambiguity ($A$), prompt template ($B$), and in-context examples ($E$), conditioned on input $X$. Assume that the LLM output $Y$ is causally dependent on this variable. Such a situation could be depicted with a Bayes net as $\psibayesnet$. \textit{We acknowledge that while this setting is oversimplistic, unlikely to occur in the real world, and do not lead to uncertainty reduction, it should serve as a starting point to think about jointly managing sources of uncertainty under the BME framework.}
        
        In this setup, the conditional independence of $A$, $B$, and $E$ given $X$ gives us:

        \begin{equation}
            I(Y;(A, B, E)|X) = I(Y;A|X) + I(Y;B|X) + I(Y;E|X)
        \end{equation}

        When the conditional independence does not hold, however, the data processing inequality~\citep{cover_elements_2012} still yields
        
        \begin{align}
            I(Y;(A, B, E)|X) \leq I[Y;\Psi|X]
        \end{align}

        While such an upper bound \textit{should not be used for maximization} to actively reduce uncertainty, an upper bound for reducible uncertainty implies a lower bound for irreducible uncertainty, as shown by \citet{yadkori_believe_2024}. Since $A$, $B$, and $E$ represent the only actionable uncertainty sources we are aware of, the residual of uncertainty -- i.e., the portion of uncertainty that cannot be reduced through manipulating these components -- can be treated as effectively irreducible or effectively aleatoric w.r.t $(A, B, E)$. This fact entails
        
        \begin{align}
            AU_{A, B, E} &= H[P(Y|X)] - I[Y;(A, B, E)|X] \\
                         &\geq H[P(Y|X)] - I[Y; \Psi|X]
        \end{align}

        In this case, such a lower bound for aleatoric uncertainty elicits a grounded notion of \textit{effective aleatoric uncertainty} with respect to \textit{known sources of epistemic uncertainty}. 
        
        This lower bound, even if being loose, remains useful in applications requiring \textit{conservative} decision-making (e.g., abstention). It represents a \textit{pragmatic best-case scenario} for aleatoric uncertainty grounded with the specified sources. Thus, it provides a grounded rationale for conservative fallback strategies, such as abstention, when the bound remains high.
        
        This line of inquiry opens up a promising research agenda: How can we design or approximate such unifying latent variables in real-world systems? What trade-offs exist between interpretability, tractability, and uncertainty reduction? And to what extent can we operationalize these theoretical bounds to inform robust deployment decisions in practice?        

    \subsection{Identifying or Rejecting Sources of Uncertainty \textit{A Priori}}

        One drawback of quantifying different sources of uncertainty using BME is that the sources must be hypothesized \textit{a priori}. BME still requires a set of experimental configurations representing each candidate source, such as unlabeled data for active learning, or predefined prompt templates for prompt engineering. Obtaining such configurations could be costly, such as acquiring user clarification of explicit intents or web-retrieved documents. While high-quality predictive simulators can sometimes alleviate this burden, as demonstrated for clarifications of intent~\citep{kobalczyk_active_2025}, such simulators are not always available or reliable.
        
        Therefore, the ability to identify or reject potential sources of uncertainty \textit{before} committing resources to collect the necessary data or stimuli is highly valuable. It helps to avoid wasted effort on factors that do not significantly contribute to uncertainty, enabling more efficient experimental design and resource allocation.

        Perhaps it would be fruitful to learn and draw inspiration from mature communities such as fractional factorial design~\citep{montgomery_design_2017}, a field of statistical experimental design that systematically reduces the number of experimental runs needed by testing only a carefully chosen subset (or fraction) of factors. Notice, however, while it's possible to learn from the principles of such fields, their technical tools might not be applicable to settings of LLM deployments. Fractional factorial design, for example, only handles factors with low-support variables (typically binary or ternary variables), and cannot be directly applied to LLMs with sequence-valued variables.
        
    \subsection{Design Experiments to Manage Uncertainty from Latent Sources}
    
        So far, the examples demonstrate that we can measure and reduce uncertainty from a particular source had we had experiments $X$ representing that source. The power of the BME framework, however, does not stop at answering only direct hypotheses. We refer here to an example of a creative experiment where BME might be used to \textit{detect the lack of in-weight knowledge} in LLMs.

         \citet{du_context_2024} investigate the interplay between in-weight knowledge and external context provided in the prompt and its effect on the answer distribution, particularly to factoid questions about \textit{entities} -- e.g., \verb|What is the capital of {|$e$\verb|}?|. They hypothesize that LLMs do not behave identically for different contexts and entities, thus employ the information gain $H[P(Y|e)] - H[P(Y|e, C)]$ to quantify the ``susceptibility'' of the LLM to questions about a fixed entity $e$. 
         
         Under the BME view, the information gain from experimenting with contexts $C$ (potentially in conflicts with pretraining data) quantifies the uncertainty reducible by contexts, \textit{indicating the lack of innate certainty from in-weight knowledge of the model} about a fixed entity $e$. A high information gain indicates the model is highly influenced by the context, where the upper bound $H[P(Y|e)]$ designates the extreme where $H[P(Y|e, C)] = 0$, implying the context fully determines the answer by the LLM. Thus, entities with \textit{maximal} information gain indicate ``shaky knowledge'' in the weights of LLMs.
         
         Alternatively, \citet{du_context_2024} also demonstrate that by fixing a context $c$ in the prompt and experimenting with different entities, we can measure the ``persuasiveness'' of the prompt across different entities $e$ under a fixed LLM -- \verb|persuasiveness(|$c$\verb|)|$= H[P(Y|E)] - H[P(Y|E, c)]$. Although this persuasiveness measure is not a type of uncertainty, it shows the power of BME in potentially eliciting interesting research questions simply by modifying different components of the experiment apparatus.

         These examples demonstrate that adopting the BME view could potentially enable researchers to think about indirectly ``interrogating'' different qualities of LLMs through appropriate designs of experiments.

\section{Conclusion and Alternative Views}\label{sec:conclusion}
        \begin{table}[h]
        \small
            \centering
            \begin{tabular}{ccc}
                \toprule
                 & \textbf{Knowns} & \textbf{Unknowns}\\
                 \midrule
                \textbf{Known} & Deterministic (\textit{Known Knowns}) & Risk (\textit{Known Unknowns}) \\[0.5em]
               \textbf{ Unknown}  & Implicit Knowledge (\textit{Unknown Knowns}) & Knightian Uncertainty (\textit{Unknown Unknowns})\\
               \bottomrule
            \end{tabular}
            \hfill \vspace{5pt}
            \caption{The quadrants of knowledge \citep{lehman_evolution_2025}.
            }
            \label{tab:quad}
            \vspace{-1.5em}
        \end{table}

    In this position paper, we argue that Bayesian Modeling of Experiments can provide a valuable framework for improving the reliability of LLMs. BME aligns with and extends existing methods aimed at quantifying and reducing uncertainty in LLMs, and is consistent with conventional notions of uncertainty in machine learning (see Appendix~\ref{sec:ml-retro}). We believe that the coherence, interpretability, and compatibility of BME for both quantification and optimization hold great potential in managing uncertainty arising from different sources in LLM generation. 

    There are \textit{alternative views} regarding our position. A typical criticism against Bayesian quantification of risk and uncertainty is that it requires defining probability distributions over all possible outcomes, which is traditionally only applicable to closed-world settings. As such, it prohibits Bayesian methods from dealing with Knightian uncertainty, i.e., uncertainty from the unknown unknowns~\citep{lehman_evolution_2025}. Yet, a probability distribution over the infinite support, such as a language model over all possible sequences, is able to represent the unknown unknowns by subsuming them into rare events or \textit{the missing mass}. This modeling assumption effectively transforms the setting of the third quadrant in Table~\ref{tab:quad} into the fourth quadrant of unknown knowns. The limit of Bayesian modeling in LLMs, hence, might not be the same as with traditional Bayesian modeling and warrants exciting future investigation, especially given the available tools in dealing with the missing mass~\citep{berend_concentration_2013, kalai_calibrated_2024, yadkori_believe_2024}.

    Another common critique against Bayesian methods is the need to assume an arbitrary prior, which is often chosen for convenience in derivations. We highlight for some applications of BME in LLMs we can bypass the need of an arbitrary prior when applied on sources that could be simulated by the LLMs -- e.g., clarification~\citep{kobalczyk_active_2025} -- allowing directly sampling from its predictive distribution. These factors, in general, could be loosely interpreted as ``within the knowledge of LLMs''. We acknowledge that the assumption of priors for factors beyond the knowledge of the LLM remains problematic and warrants further research, especially for the current novel infinite-support regime.

    Finally, some researchers may contend that reliability issues will naturally diminish as language models scale, making dedicated research on managing uncertainty potentially redundant. Indeed, large language models continue to improve in both capabilities and calibration, as shown in benchmark evaluations~\citep[\textit{inter alia}]{srivastava_beyond_2023, chiang_chatbot_2024, phan_humanitys_2025}. However, uncertainty is an inherent feature of natural language interactions and open-ended environments. On top of that, there will always be ``tail'' domains where textual training data is scarce, e.g., tacit knowledge -- knowledge that is hard to codify explicitly~\citep{kambhampati_polanyis_2021}, leading to inevitable uncertainty in reasoning.  Consequently, research into uncertainty, particularly in identifying its sources and developing methods to mitigate it, will remain crucial, even with the scaling of LLMs.

\section*{Acknowledgement}
    We thank Martin Strobel and Rishav Chourasia for their valuable discussions that led to the conception of this work and for their feedback on this manuscript.
    This research is supported by the Ministry of Digital Development and Information (MDDI) under the Singapore Global AI Visiting Professorship Program (Award No. AIVP-2024-002), and the National Research Foundation, Singapore under its AI Singapore Programme (AISG Award No:  AISG3-PhD 2023-08-053).

\bibliography{references/zotero}
\bibliographystyle{jae}

\newpage
\onecolumn
\appendix
\section*{APPENDIX}

\section{What Is Reducible Depends on the Experiment Apparatus}\label{sec:ml-retro}        
    In this section, we show that \textbf{the BME framework is compatible with classical distinctions in the machine learning literature}, in which epistemic uncertainty pertains to model parameters, while aleatoric uncertainty pertains to inherent stochasticity in data~\citep{ling_uncertainty_2024, hou_decomposing_2024, smith_rethinking_2024, gal_uncertainty_2016, kendall_what_2017}. Moreover, not only does it subsumes these notions within a coherent and principled framework, but it also elucidates the reasoning behind their use. 

    \subsection{Uncertainty Quantification in Machine Learning Under Adaptive Data Acquisition Setting}

    The most popular application of Bayesian modeling of experiments is the application of Bayesian Experimental Design in \textit{active learning}, whose goal is to select the most informative unlabeled data to be annotated and acquired into the training dataset to improve the learned parametric model $\theta$.

     Bayesian Experimental Design (BED)~\citep{rainforth_modern_2024} more specifically distinguishes components of an experiment $\Lambda$ into its design $X$ and outcome (unknown before performing the experiment) $Y$, i.e., an experiment $\lambda$ is representable as a tuple $(x, y)$. It is then customary to assume $Y$ and $\Xi$ together specify $P(Y|\Xi)$, while $\Xi \perp X$ and $P(\Xi|X) = P(\Xi)$.

     As before, the pointwise information gain from an experiment is the difference in uncertainty going from the prior $P(\Xi)$ to the posterior $P(\Xi| x, y)$

     \begin{align}
         I[\Xi; x,y] &= H[P(\Xi)] - H[P(\Xi | x, y)].
     \end{align}

    The focus of BED, however, is estimating the expected information gained from a design $x$, across all possible outcomes $y|x$. To this, modern BED often construct a predictive model (i.e., a simulator) $P(Y|X)$ for possible experiment outcomes $y$, given an experiment configuration $x$. The expected information gain is defined as the expectation taken over the simulator $P(Y|x)$.
     
    \begin{align}
        EIG_\Xi(x) \coloneqq &\mathbb{E}_{P(Y | x)} \left[ H[P(\Xi)] - H[P(\Xi | x, y)] \right]\\
                             &= H[P[\Xi]] - \mathbb{E}_{P(Y | x)} H[P(\Xi | x, y)] \\
                                &= D_{KL}(P(\Xi, Y| x) \| P(\Xi) P(Y|x))\\
                                &= I(\Xi; Y|x)
    \end{align}

    Note that we still recover the KL-divergence similar to before, thanks to the independence assumption $\Xi \perp X$. However, the quantity in this case is the conditional mutual information, which doesn't tell you the information gain about $\Xi$ after an experiment, but rather the \textit{potential information gain} about $\Xi$ according to your simulator $P(Y|X)$ had you chosen the design $x$. Thus, it is often used as an objective for optimization over design $X$.

    In \textit{active learning}, we take the parameter $\theta$ of the parametric model (e.g., a neural network) as the quantity of interest. The experiment stimulus, in this case, is the acquisition of unlabeled data points $(x, \varnothing)$ into the training dataset.
    
    \citet{houlsby_bayesian_2011} correspondingly propose the \textit{Bayesian Active Learning by Disagreement} (BALD) score as the expected information gain from labeling $x$.

    \begin{align}
        BALD(x) &= H[P(\Theta)] - \mathbb{E}_{P(Y | x)} H[P(\Theta | x, y)]\\
                &= I(\Theta; Y|x)\\
                &= H[P(Y|x)] - \mathbb{E}_{P(\Theta)} H[P(Y|x, \theta)] \label{eq:flip-bald}
    \end{align}

    Equation~\ref{eq:flip-bald} exploits the symmetry of the mutual information to enable efficient computation of $BALD$, requiring only taking expectation over $P(\Theta)$ by Monte-Carlo approximation over ensembles, instead of dealing explicitly with the high-dimensional probability space. 

    In this setting of \textit{active learning}, epistemic uncertainty is seen as \textit{data uncertainty}, which contradicts the convention in machine learning in the fixed data setting below. This inconsistency in conventions is known to cause confusion in communication between researchers from the two regimes.
     
    \subsection{Uncertainty Quantification in Classical Machine Learning Under Fixed Data Setting}
        In prior machine learning literature, tasks are typically defined with a fixed set of data, making aleatoric uncertainty, stemming from noise or variability in the data, synonymous with \textit{data uncertainty}. Similarly, epistemic or reducible uncertainty is typically attributed to the divergence between the learned parameterized model and the true data-generating process~\citep{fong_martingale_2023}.

        In a typical machine learning problem, we have access to a finite set of training data $d_{1:n} \sim P(X,Y)$. Denote $P_{n}(Y|X) \triangleq P(Y|X, d_{1:n})$ as the model learned from the finite dataset, and assume $P_n(Y|X) \to P_\infty(Y|X)$ as $n \to \infty$, where $P_\infty(Y|X)$ is the true data generating process. Further assume that the true data-generating process is parameterized by a parameter $\theta^*$ -- i.e., $P(Y|X, \theta^*) \coloneqq P_\infty(Y|X)$.
        
        Under the Bayesian perspective, given a data-generating process and a consistent Bayesian learner model, the reduction in epistemic uncertainty $EU$ is due to data that has not yet been observed~\citep{fong_martingale_2023, wang_subjective_2024, smith_rethinking_2024}, i.e., 

        \begin{equation}
             EU \coloneqq H[P_n(Y|X)] - H[P_\infty(Y|X)]
        \end{equation}

        Since we do not have infinite data in practice, $H[P_\infty(Y|X)]$ is approximated by its Bayes estimator $\mathbb{E}_{P_n(\Theta)}[H[P(Y|\theta, X)]]$ under a quadratic loss $\ell(h) = (h - H[P_\infty(Y|X)])^2 = (h - H[P(Y|X, \theta^*)])^2$~\citep{smith_rethinking_2024}(The proof is provided in Appendix~\ref{sec:bayes-est}).

        \begin{align}
            EU &\coloneqq H[P_n(Y|X)] - H[P_\infty(Y|X)]\\
               &\approx H[P_n(Y|X)] - \mathbb{E}_{P_n(\Theta)}[H[P(Y| X, \theta)]] \label{eq:gal-approx}\\
               &= \mathbb{E}_{x \sim P_n(X)} D_{KL}(P_n(Y, \Theta | x) \| P_n(Y|x) P_n(\Theta))\\
               &= \mathbb{E}_{x \sim P_n(X)} EIG(x) \\
               &= I[Y|X; \Theta]
        \end{align}

        As such, the quantification of epistemic uncertainty coincides with the BME framework, where the quantity of interest is $Y|X$ (not with respect to any specific $x$) and experiments are models of different $\theta$ learned via training data points $d_{1:n}$.~\footnote{The approximated formula \ref{eq:gal-approx} is actually first proposed by \citet{gal_uncertainty_2016, kendall_what_2017}, based on inspiration from the BALD score and the intuition that aleatory is ``the noise inherent in the observations'' persisting across models. We present above a first-principle derivation of the formula.}

        It is obvious that under this setting, epistemic uncertainty was solely attributed to \textit{model uncertainty}, or \textit{parameter uncertainty} as it was the only `` tuning knob'' to perform experiments given fixed data. The remaining residual uncertainty, estimated by $\mathbb{E}_{P_n(\Theta)}[H[P(Y| X, \theta)]]$ is often called both \textit{aleatoric uncertainty} and \textit{data uncertainty}. Though confusing to researchers in active learning, these are more common conventions, due to \citet{gal_uncertainty_2016, kendall_what_2017, abdar_review_2021} and persist in newer research in LLMs~\citep{hou_decomposing_2024, ling_uncertainty_2024}.
                
\section{The Bayes-optimal Estimator of Irreducible Predictive Entropy}\label{sec:bayes-est}

\begin{proposition}[\citep{smith_rethinking_2024}]
    A model's conditional predictive entropy $\mathbb{E}_{P_n(\Theta)} H[P(Y|X, \theta)]$ is a Bayes estimator of its irreducible predictive entropy $H[P_\infty(Y|X)]$
\end{proposition}
\begin{proof}
    (Adapted from~\citep{smith_rethinking_2024})
    
    Let $h$ be an estimator of $H[P_\infty(Y|X)]$, and let $P_n(\Theta)$ be our beliefs about which $\theta$ is $\theta^*$ that yield $P(Y|X, \theta^*) = P_\infty(Y|X)$.

    Consider the quadratic loss with respect to the irreducible predictive entropy $\ell(h, H[P_\infty(Y|X)]) = (h - H[P(Y|X, \theta^*)])^2$. 
    
    By definition, the Bayes estimator is the minimizer of the expected loss over the posterior expected loss (Bayes risk)
    \begin{align}
        L(h) &= \mathbb{E}_{P_n(\Theta)}[ \ell(h, \theta)]\\
             &= \mathbb{E}_{P_n(\Theta)}[(h - H[P(Y|X, \theta)])^2]
    \end{align}

    The Bayes-optimal estimator $h^*$ satisfies
    \begin{equation}
        \nabla_{h^*} L(h^*) = \mathbb{E}_{P_n(\Theta)}[2(h^* - H[P(Y|X, \theta)])] = 0,
    \end{equation}

     which gives $h^* = \mathbb{E}_{P_n(\Theta)}[H[P(Y|X, \theta)])]$.   
    
\end{proof}

\section{Monte-Carlo Estimator of the Marginal Entropy is a Variance-Corrected Estimator of the Conditional Entropy}\label{sec:taylor-approx}
\begin{proposition} 

Let $x_1, \dots, x_N \sim p(x)$ i.i.d., and define
\begin{equation*}
    \bar{p}(y) = \frac{1}{N} \sum_{j=1}^N p(y|x_j).
\end{equation*}
Then
\begin{equation*}
    \sum_y \bar{p}(y) \log \bar{p}(y) \approx -H[P(Y|X)] - \frac{1}{2} \sum_y \frac{\operatorname{Var}_j(p(y|x_j))}{\bar{p}(y)},
\end{equation*}

where $\operatorname{Var}_j(p(y|x_j))$ denotes the sample variance of $p(y|x_j)$ over $j = 1, \dots, N$.
\end{proposition}

\begin{proof}
Define
\begin{equation}
    f(q) = q \log q,
\end{equation}
which is convex for $q > 0$. We can write:
\begin{equation}
    A := \sum_y \bar{p}(y) \log \bar{p}(y) = \sum_y f(\bar{p}(y)).
\end{equation}
Define the Monte Carlo estimate of the conditional entropy:
\begin{equation}
    B_N := \frac{1}{N} \sum_{j=1}^N \sum_y p(y|x_j) \log p(y|x_j) = \frac{1}{N} \sum_{j=1}^N \sum_y f(p(y|x_j)).
\end{equation}
For each $y$, define $\delta_{yj} = p(y|x_j) - \bar{p}(y)$ so that
\begin{equation}
    \frac{1}{N} \sum_{j=1}^N \delta_{yj} = 0.
\end{equation}
Apply a second-order Taylor expansion of $f$ around $\bar{p}(y)$:
\begin{equation}
    f(p(y|x_j)) \approx f(\bar{p}(y)) + f'(\bar{p}(y)) \delta_{yj} + \frac{1}{2} f''(\bar{p}(y)) \delta_{yj}^2,
\end{equation}
where
\begin{equation}
    f'(q) = \log q + 1, \quad f''(q) = \frac{1}{q}.
\end{equation}
Substitute into $B_N$:
\begin{align}
    B_N &\approx \frac{1}{N} \sum_{j=1}^N \sum_y \left[ f(\bar{p}(y)) + (\log \bar{p}(y) + 1)\delta_{yj} + \frac{1}{2\bar{p}(y)}\delta_{yj}^2 \right] \\
    &= \sum_y f(\bar{p}(y)) + \sum_y (\log \bar{p}(y) + 1)\left( \frac{1}{N} \sum_{j=1}^N \delta_{yj} \right) + \sum_y \frac{1}{2\bar{p}(y)} \cdot \frac{1}{N} \sum_{j=1}^N \delta_{yj}^2.
\end{align}
The second term vanishes by construction:
\begin{equation}
    \frac{1}{N} \sum_{j=1}^N \delta_{yj} = 0.
\end{equation}
Thus,
\begin{equation}
    B_N \approx A + \sum_y \frac{\operatorname{Var}_j(p(y|x_j))}{2\bar{p}(y)}.
\end{equation}
Rearranging gives:
\begin{equation}
    A \approx B_N - \frac{1}{2} \sum_y \frac{\operatorname{Var}_j(p(y|x_j))}{\bar{p}(y)}.
\end{equation}
Finally, since $B_N$ approximates $B = -H[P(Y|X)]$ by the Law of Large Numbers, we conclude:
\begin{equation}
    \sum_y \bar{p}(y) \log \bar{p}(y) \approx -H[P(Y|X)] - \frac{1}{2} \sum_y \frac{\operatorname{Var}_j(p(y|x_j))}{\bar{p}(y)}.
\end{equation}
\end{proof}

\section{A Formal Model of Language Model Generation}~\label{sec:formal-lm}
    Let $\mathcal{V}$ be the token vocabulary, and the Kleene closure $\mathcal{V}^*$ is the set of all strings composable from elements of $\mathcal{V}$. Consider a language model $f_{LM}$ as a probability mass function over the Kleene closure $\mathcal{V}^*$ that maps a sequence of tokens to a probability vector over the Kleene closure $\mathcal{V}$ i.e., $f_{LM}: \mathcal{V}^* \to \Delta^{|\mathcal{V}^*|}$. We denote $P_{LM}$ as the probability distribution characterized by $f_{LM}$. 
    
    Note that the abstraction above diverges from the actual construction of a generative language model as an autoregressive (next-token) predictor. This implicitly assumes that, for a variable-length sequence of input tokens $x = \{x_1, \dots, x_T\}$ and $y = \{y_1, \dots, y_L\}$, the mass of the generating output sequences $p_{LM}(y \mid x) \coloneqq [f_{LM}(x)]_{y}$ is recursively factorizable according to the Markov assumption

    \begin{equation}
       p_{LM}(y \mid x) = \prod_{l = 1}^{L} p_{AR}(y_l \mid y_{< l}, x),
        \label{eq:factorized-seq}
    \end{equation}

    where $P_{AR}$ is the probability distribution characterized by the actual pretrained next-token predictor $f_{AR}: \mathcal{V}^* \to \Delta^{|\mathcal{V}|}$.

\end{document}